\documentclass[11pt,a4paper]{article}
\usepackage[hyperref]{acl2019}
\aclfinalcopy
\usepackage{times}
\usepackage{latexsym}

\usepackage{url,flushend}
\usepackage[utf8]{inputenc}
\title{Cognitive Graph for Multi-Hop Reading Comprehension at Scale}

\author{Ming Ding\textsuperscript{\textdagger}, Chang Zhou\textsuperscript{$\ddagger$}, Qibin Chen\textsuperscript{\textdagger}, Hongxia Yang\textsuperscript{$\ddagger$}, Jie Tang\textsuperscript{\textdagger} \\
  \textsuperscript{\textdagger}Department of Computer Science and Technology, Tsinghua University \\
  \textsuperscript{$\ddagger$}DAMO Academy, Alibaba Group \\
  \texttt{\{dm18,chen-qb15\}@mails.tsinghua.edu.cn} \\
  \texttt{\{ericzhou.zc,yang.yhx\}@alibaba-inc.com}\\
  \texttt{jietang@tsinghua.edu.cn}
  }
\date{\today}
\usepackage{graphicx}
\usepackage[linesnumbered,boxed,ruled,commentsnumbered]{algorithm2e}
\usepackage{amsmath}
\usepackage{multirow}
\usepackage{titlesec}
\usepackage{amsfonts}
\usepackage{paralist}
\def\*#1{\mathbf{#1}}
\def\&#1{\mathcal{#1}}
\DeclareMathOperator*{\argmax}{arg\,max}

\usepackage{xcolor}
\definecolor{comment}{RGB}{70, 150, 60}

\newcommand{\eat}[1]{}

\newcommand{\sysone}{System 1}
\newcommand{\systwo}{System 2}
\newcommand{\name}{CogQA}
\newcommand{\vpara}[1]{\vspace{0.05in}\noindent\textbf{#1 }}
\begin{document}
\addtolength{\abovedisplayskip}{-5pt}
\addtolength{\belowdisplayskip}{-6pt}
\maketitle
\begin{abstract}
We propose a new \name~framework for multi-hop 
question answering in web-scale documents. Founded on the dual process theory in cognitive science, the framework gradually builds a \textit{cognitive graph} in an iterative process by coordinating an \textbf{implicit extraction} module (System 1) and an \textbf{explicit reasoning} module (System 2). While giving accurate answers, our framework further provides explainable reasoning paths. Specifically, our implementation\footnote{Codes are avaliable at \url{https://github.com/THUDM/CogQA}} based on BERT and graph neural network (GNN) efficiently handles millions of documents for multi-hop reasoning questions in the HotpotQA fullwiki dataset, achieving a winning joint $F_1$ score of 34.9 on the leaderboard, compared to 23.6 of the best competitor.\footnote{\url{https://hotpotqa.github.io}, March 4, 2019}
\end{abstract}
\section{Introduction}\label{sec:intro}
Deep learning models have made significant strides in machine reading comprehension and even outperformed human on single paragraph question answering (QA) benchmarks including SQuAD~\citep{wang2018multi,devlin2018bert,rajpurkar2016squad}. However, to cross the chasm of reading comprehension ability between machine and human, three main challenges lie ahead:
\textbf{1)} Reasoning ability. As revealed by adversarial tests~\citep{jia2017adversarial}, models for single paragraph QA tend to seek answers in sentences matched by the question, which does not involve complex reasoning. Therefore, multi-hop QA becomes the next frontier to conquer~\cite{yang2018hotpotqa}. \textbf{2)} Explainability. 
Explicit reasoning paths, which enable verification of logical rigor, are vital for the reliability of QA systems. HotpotQA~\cite{yang2018hotpotqa} requires models to provide supporting sentences, which means \textit{unordered} and \textit{sentence-level} explainability, yet humans can interpret answers with step by step solutions, indicating an \textit{ordered} and \textit{entity-level} explainability. \textbf{3)} Scalability. For any practically useful QA system, scalability is indispensable. Existing QA systems based on machine comprehension generally follow retrieval-extraction framework in DrQA~\cite{chen2017reading}, reducing the scope of sources to a few paragraphs by pre-retrieval. This framework is a simple compromise between single paragraph QA and scalable information retrieval, compared to human's ability to breeze through reasoning with knowledge in massive-capacity memory~\cite{wang2003discovering}. 
\begin{figure}
    \centering
    \includegraphics[width=\linewidth]{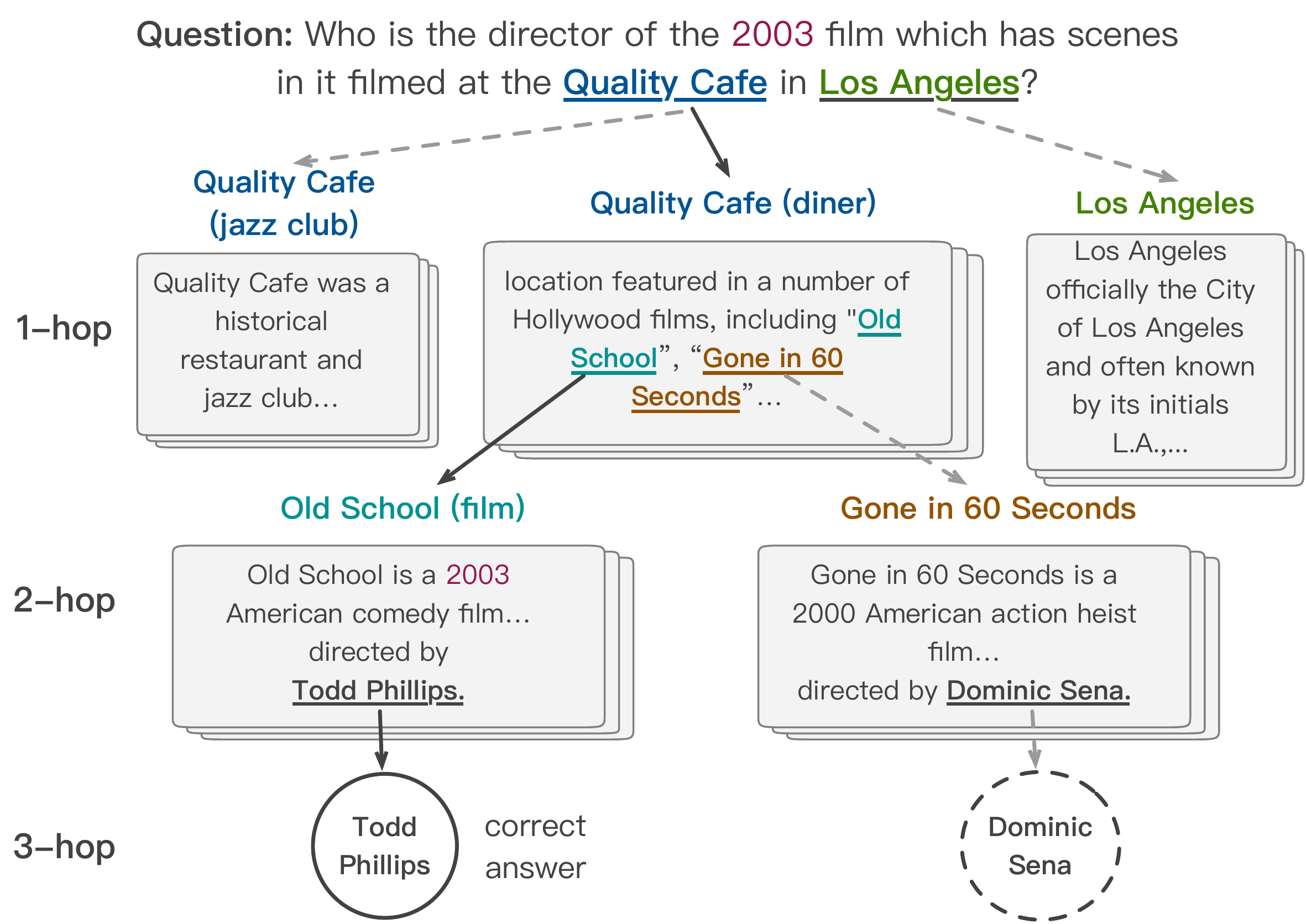}
    \setlength{\abovecaptionskip}{0pt}
    \setlength{\belowcaptionskip}{-15pt}
    \caption{An example of cognitive graph for multi-hop QA. Each \textit{hop node} corresponds to an entity (e.g., ``Los Angeles'') followed by its introductory paragraph. The circles mean \textit{ans nodes}, answer candidates to the question. Cognitive graph mimics human reasoning process. Edges are built when calling an entity to ``mind''. The solid black edges are the correct reasoning path.}    \label{fig:example}
\end{figure}

Therefore, insights on the solutions to these challenges can be drawn from the cognitive process of humans.
\emph{Dual process theory}~\cite{evans1984heuristic,evans2003two,evans2008dualreview,sloman1996empirical} suggests that our brains first retrieve relevant information following attention via an \textit{implicit}, \textit{unconscious} and \textit{intuitive} process called \textbf{System 1}, based on which another \textit{explicit}, \textit{conscious} and \textit{controllable} reasoning process, \textbf{System 2}, is then conducted. 
System 1 
could 
provide resources according to requests, while System 2 enables diving deeper into relational information by performing sequential thinking in the working memory, which is slower but with human-unique rationality~\citep{baddeley1992working}. For complex reasoning, the two systems are coordinated to perform \textit{fast and slow thinking}~\cite{kahneman2011thinking} iteratively. 

In this paper, we propose a framework, namely \emph{Cognitive Graph QA} (\name), contributing to tackling all challenges above. Inspired by the dual process theory, the framework comprises functionally different System 1 and 2 modules.
System 1 extracts question-relevant entities and answer candidates from paragraphs and encodes their semantic information. Extracted entities are organized as a \emph{cognitive graph} (Figure~\ref{fig:example}), which resembles the working memory. System 2 then conducts the reasoning procedure over the graph, and collects \textit{clues} to guide System 1 to better extract next-hop entities. The above process is iterated until all possible answers are found, and then the final answer is chosen based on reasoning results from System 2.
An efficient implementation based on BERT~\cite{devlin2018bert} and graph neural network (GNN)~\cite{battaglia2018relational} is introduced. 

Our contributions are as follows:
\begin{itemize}
\setlength{\itemsep}{-3pt}
\setlength{\parsep}{0pt}
\setlength{\parskip}{-10pt}
    \item We propose the novel \name~framework for multi-hop reading comprehension QA at scale according to human cognition.\\
    \item We show that the \textit{cognitive graph} structure in our framework offers ordered and entity-level explainability and suits for relational reasoning.\\ 
    \item Our implementation based on BERT and GNN surpasses previous works and other competitors substantially on all the metrics.
\end{itemize}


 

\section{Cognitive Graph QA Framework}\label{frame}
\begin{algorithm}[hbt]
\small
\SetInd{1pt}{7pt}
\caption{Cognitive Graph QA}\label{algo}
\KwIn{
    \\
    System 1 model $\&S_1$, System 2 model $\&S_2$,\\
    Question $Q$, Predictor $\&F$,Wiki Database $\&W$\\
    }
Initialize cognitive graph $\&G$ with entities mentioned in $Q$ and mark them \emph{frontier nodes}\\
\Repeat{there is no frontier node in $\&G$ or $\&G$ is large enough}{
    pop a node $x$ from frontier nodes\\
    collect $clues[x,\&G]$ from predecessor nodes of $x$\ \ \ \ \ \ \ \ {\color{comment}// eg. $clues$ can be sentences where $x$ is mentioned}\\
    fetch $para[x]$ in $\&W$ if any\\
    generate $sem[x,Q,clues]$ with $\&S_1$ {\color{comment}// initial $\*X[x]$}\\
    \If{$x$ is a hop node}{
        find hop and answer spans in $para[x]$ with $\&S_1$\\
        \For{$y$ in hop spans}{
        \uIf{$y\notin \&G$ and $y \in \&W$}{create a new hop node for $y$}
        \uIf{$y \in \&G$ and edge$(x,y)\notin \&G$}{add edge $(x,y)$ to $\&G$\\ mark node $y$ as a frontier node}
        }
        \For{$y$ in answer spans}{
            add new answer node $y$ and edge $(x,y)$ to $\&G$
        }
    }
    update hidden representation $\mathbf{X}$ with $\&S_2$
}
\textbf{Return} $\argmax\limits_{\text{answer node }x} \&F(\*X[x])$
\end{algorithm}
\begin{figure*}[ht]
    \centering
    \setlength{\abovecaptionskip}{5pt}
    \setlength{\belowcaptionskip}{-10pt}
    \includegraphics[width=0.9\linewidth]{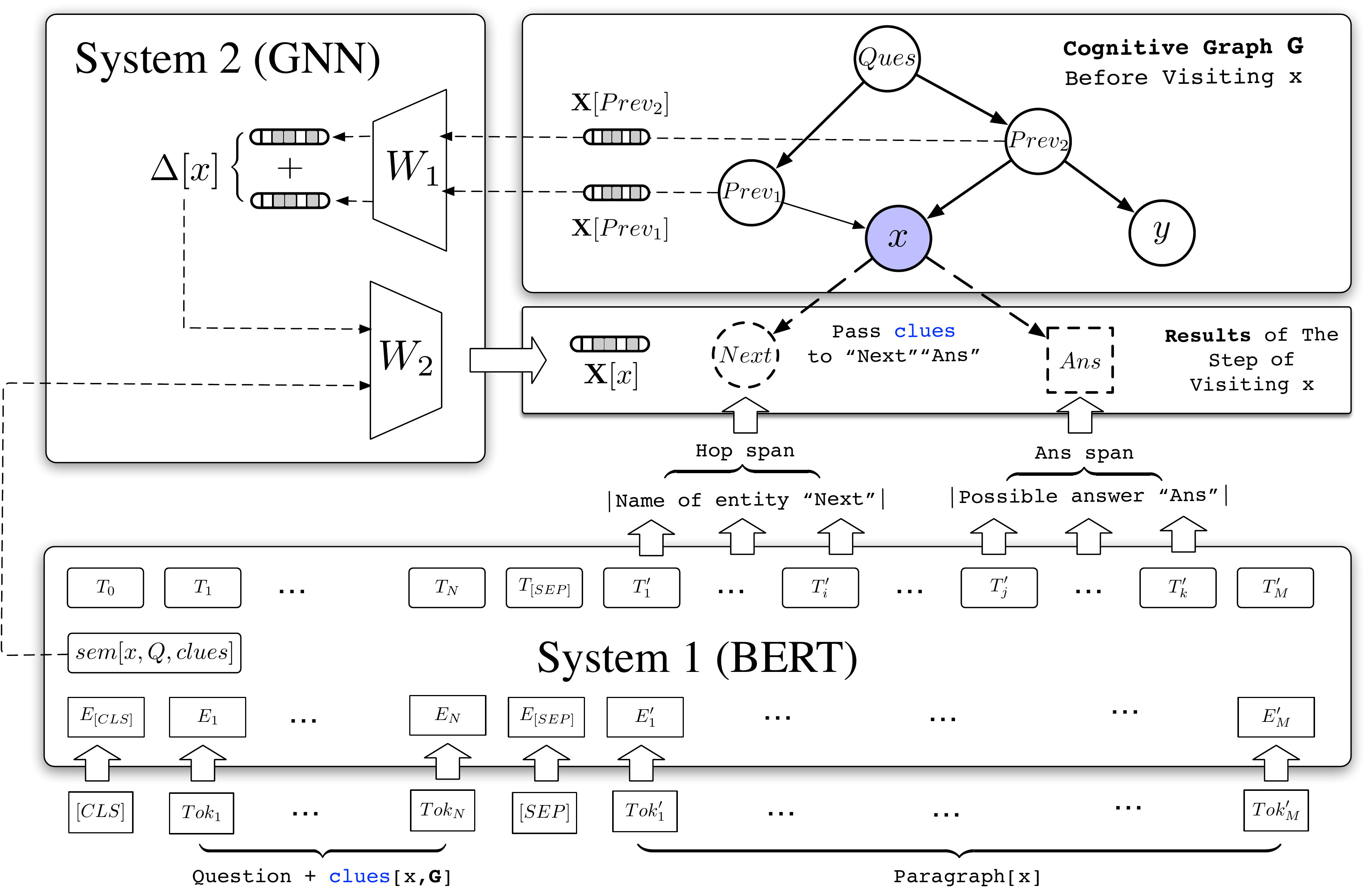}
    \caption{Overview of \name~implementation. When visiting the node $x$, \sysone~generates new hop and answer nodes based on the $clues[x,\&G]$ discovered by \systwo.
    It also creates the inital representation $sem[x,Q,clues]$, based on which 
    the GNN in \systwo~updates the hidden representations $\mathbf{X}[x]$.
    }
    \label{fig:model}
\end{figure*}
Reasoning ability of humankind depends critically on relational structures of information. Intuitively, we adopt a directed graph structure for step-by-step deduction and exploration in cognitive process of multi-hop QA.
In our reading comprehension setting, each node in this \emph{cognitive graph} $\&G$ corresponds with an entity or possible answer $x$, also interchangeably denoted as node $x$. The extraction module System 1, reads the introductory paragraph $para[x]$ of entity $x$ and extracts \textit{answer candidates} and useful \textit{next-hop entities} from the paragraph. $\&G$ is then expanded with these new nodes, providing explicit structure for the reasoning module, System 2. In this paper, we assume that System 2 conducts deep learning based instead of rule-based reasoning by computing hidden representations $\*X$ of nodes. Thus System 1 is also required to summarize $para[x]$ into a semantic vector as initial hidden representation when extracting spans. Then System 2 updates $\*X$ based on graph structure as reasoning results for downstream prediction. 

Explainability is enjoyed owing to explicit reasoning paths in the cognitive graph. Besides simple paths, the cognitive graph can also clearly display joint or loopy reasoning processes, where new predecessors might bring new \textit{clues} about the answer. \textit{Clues} in our framework is a form-flexible concept, referring to information from predecessors for guiding System 1 to better extract spans. Apart from newly added nodes, those nodes with new incoming edges also need revisits due to new clues. We refer to both of them as \textit{frontier nodes}. 

Scalability means that the time consumption of QA will not grow significantly along with the number of paragraphs. Our framework can scale in nature since the only operation referred to all paragraphs is to access some specific paragraphs by their title indexes. For multi-hop questions, traditional retrieval-extraction frameworks might sacrifice the potential of follow-up models, because paragraphs multiple hops away from the question could share few common words and little semantic relation with the question, leading to a failed retrieval. However, these paragraphs can be discovered by iteratively expanding with $clues$ in our framework.

Algorithm~\ref{algo} describes the procedure of our framework \name. After initialization, an iterative process for graph expansion and reasoning begins. In each step we visit a frontier node $x$, and System 1 reads $para[x]$ under the guidance of $clues$ and the question $Q$, extracts spans and generates semantic vector $sem[x,Q,clues]$. Meanwhile, System 2 updates hidden representation $\*X$ and prepares $clues[y,\&G]$ for any successor node $y$. The final prediction is made based on $\*X$.

\section{Implementation}\label{sec:implement}
The main part to implement the \name~framework is to determine the concrete models of System 1 and 2, and the form of $clues$. 

Our implementation uses BERT as System 1 and GNN as System 2. Meanwhile, $clues[x,\&G]$ are sentences in paragraphs of $x$’s predecessor nodes, from which $x$ is extracted.
We directly pass raw sentences as $clues$, rather than any form of computed hidden states, for easy training of \sysone.
Because raw sentences are self-contained and independent of computations from previous iterative steps, training at different iterative steps is then decoupled, leading to efficiency gains during training. Details are introduced in \S~\ref{sec:train}. Hidden representations  $\mathbf{X}$ for graph nodes are updated each time by a propagation step of GNN.


Our overall model is illustrated in Figure~\ref{fig:model}.
\subsection{System 1}

The extraction capacity of \sysone~model is fundamental to construct the cognitive graph, thus a powerful model is needed. 
Recently, BERT~\citep{devlin2018bert} has become one of the most successful language representation models on various NLP tasks, including SQuAD~\citep{rajpurkar2016squad}. BERT consists of multiple layers of Transformer~\cite{vaswani2017attention}, a self-attention based architecture, and is elaborately pre-trained on large corpora. Input sentences are composed of two different functional parts A and B. 

We use BERT as \sysone, and its input when visiting the node $x$ is as follows:
\begin{equation*}
\begin{small}
\label{input}
    \underbrace{[CLS] \ Question \  [SEP] \ clues[x, \&G] \ \ [SEP]}_{Sentence\ A} \underbrace{\ Para[x]}_{Sentence\ B}
\end{small}
\end{equation*}
\noindent where $clues[x, \&G]$ are sentences passed from predecessor nodes. 
The output vectors of BERT are denoted as $\*T \in \mathbb{R}^{L\times H}$, where $L$ is the length of the input sequence and $H$ is the dimension size of the hidden representations.

It is worth noting that for answer node $x$, $Para[x]$ is probably missing. Thus we do not extract spans but can still calculate $sem[x,Q,clues]$ based on ``Sentence A'' part. And when extracting 1-hop nodes from question to initialize $\&G$, we do not calculate semantic vectors and only the $Question$ part exists in the input. 

\textbf{Span Extraction\ } Answers and next-hop entities have different properties. Answer extraction relies heavily on the character indicated by the question. For example ``New York City'' is more possible to be the answer of a \emph{where} question than ``2019'', while next-hop entities are often the entities whose description matches statements in the question. Therefore, we predict answer spans and next-hop spans separately. 

We introduce ``pointer vectors" $\*S_{hop}, \*E_{hop},$ $ \*S_{ans},$ $ \*E_{ans}$ as additional learnable  parameters to predict targeted spans. The probability of the $i^{th}$ input token to be the start of an answer span $P_{ans}^{start}[i]$ is calculated as follows:
\begin{equation}
    P_{ans}^{start}[i] = \frac { e ^ { \*S_{ans} \cdot \*T _ { i } } } { \sum _ { j } e ^ { \*S_{ans} \cdot \*T _ { j } } }
\end{equation}

Let $P_{ans}^{end}[i]$ be the probability of the $i^{th}$ input token to be the end of an answer span, which can be calculated following the same formula.
We only focus on the positions with top K start probabilities $\{start_k\}$. For each k, the end position $end_k$ is given by:
\begin{equation}
    end_k = \argmax\limits_{start_k \leq j \leq start_k + maxL} P_{ans}^{end}[j]\\
\end{equation}
\noindent where $maxL$ is the maximum possible length of spans.

To identify irrelevant paragraphs, we leverage negative sampling introduced in \S~\ref{neg} to train System 1 to generate a \textit{negative threshold}. In top K spans, those whose start probability is less than the negative threshold will be discarded.
Because the $0^{th}$ token $[CLS]$ is pre-trained to synthesize all input tokens for the Next Sentence Prediction task~\citep{devlin2018bert}, $P_{ans}^{start}[0]$ acts as the threshold in our implementation. \label{neg1}

We expand the cognitive graph with remaining predicted answer spans as new ``answer nodes''. The same process is followed to expand ``next-hop nodes'' by replacing $\*S_{ans}, \*E_{ans}$ with $\*S_{hop}, \*E_{hop}$. 

\textbf{Semantics Generation\ } As mentioned above, outputs of BERT at position 0 have the ability to summarize the sequence. Thus the most straightforward method is to use $\*T_0$ as $sem[x,Q,clues]$. However, the last few layers in BERT are mainly in charge of transforming hidden representations for span predictions. In our experiment, the usage of the third-to-last layer output at position 0 as $sem[x,Q,clues]$ performs the best.

\subsection{System 2}
The first function of System 2 is to prepare $clues[x,\&G]$ for frontier nodes, 
which we implement it as collecting the raw sentences of $x$'s predecessor nodes that mention $x$. 

The second function, to update hidden representations $\mathbf{X}$, is the core function of System 2. 
Hidden representations $\mathbf{X}\in \mathbb{R}^{n\times H}$ stand for the understandings of all $n$ entities in $\&G$. To fully understand the relation between an entity $x$ and the question $Q$, barely analyzing semantics $sem[x,Q,clues]$ is insufficient. GNN has been proposed to perform deep learning on graph~\cite{kipf2016semi}, especially relational reasoning owing to the inductive bias of graph structure~\cite{battaglia2018relational}. 

In our implementation, a variant of GNN is designed to serve as System 2. For each node $x$, the initial hidden representation $\*X[x] \in \mathbb{R}^{H}$ is the semantic vector $sem[x,Q,clues]$ from System 1. Let $\*X'$ be the new hidden representations after a propagation step of GNN, and ${\Delta} \in \mathbb{R}^{n\times H}$ be aggregated vectors passed from neighbours in the propagation. The updating formulas of $\*X$ are as follows: 
\begin{align}
    &\Delta = \sigma((AD^{-1})^T\sigma(\*XW_1))\label{eq:delta}\\
    &\*X' = \sigma(\*XW_2 + \Delta)\label{eq:gnn}
\end{align}
where $\sigma$ is the activation function and $W_1,W_2\in \mathbb{R}^{H\times H}$ are weight matrices. $A$ is the adjacent matrix of $\&G$, which is column-normalized to $AD^{-1}$ where $D_{jj}=\sum_i A_{ij}$. Transformed hidden vector $\sigma(\*XW_1)$ is left multiplied by $(AD^{-1})^T$, which can be explained as a localized spectral filter by ~\citet{defferrard2016convolutional}.

In the iterative step of visiting frontier node $x$, its hidden representation $\*X[x]$ is updated following Equation (\ref{eq:delta})(\ref{eq:gnn}). In experiments, we observe that this ``asynchronous updating'' shows no apparent difference in performance with updating $\*X$ of all the nodes together by multiple steps after $\&G$ is finalized, which is more efficient and adopted in practice.   

\subsection{Predictor}
The questions in HotpotQA dataset generally fall into three categories: \emph{special} question, \emph{alternative} question and \emph{general} question, which are treated as three different downstream prediction tasks taking $\*X$ as input. In the test set, they can also be easily categorized according to interrogative words.

Special question is the most common case, requesting to find spans such as locations, dates or entity names in paragraphs. We use a two-layer fully connected network (FCN) to serve as predictor $\&F$:
\begin{equation}
    answer = \argmax \limits_{\text{answer node }x} \&F(\*X[x])
\end{equation}

Alternative and general question both aims to compare a certain property of entity $x$ and $y$ in HotpotQA, respectively answered with entity name and ``yes or no''. These questions are regarded as binary classification with input $\*X[x] - \*X[y]$ and solved by another two identical FCNs.


\subsection{Training}\label{sec:train}
Our model is trained under a supervised paradigm with negative sampling. In the training set, the next-hop and answer spans are pre-extracted in paragraphs. More exactly, for each $para[x]$ relevant to question $Q$, we have spans data

\begin{small}
\[\&D[x,Q] = \{(y_1, start_1, end_1), ...,(y_n, start_n, end_n)\}\] 
\end{small}where the span from $start_i$ to $end_i$ in $para[x]$ is fuzzy matched with the name of an entity or answer $y_i$. See \S ~\ref{sec:dataset} for detail.

\subsubsection{Task \#1: Span Extraction}
The ground truths of $P_{ans}^{start}, P_{ans}^{end},$ $P_{hop}^{start}, P_{hop}^{end}$ are constructed based on $\&D[x, Q]$. There is at most one answer span $(y,$ $ start,$ $ end)$ in every paragraph, thus $\*{gt}_{ans}^{start}$ is an one-hot vector where $\*{gt}_{ans}^{start}[start] = 1$. However, multiple different next-hop spans might appear in one paragraph, so that $\*{gt}_{hop}^{start}[start_i] = 1/k$ where $k$ is the number of next-hop spans.  

For the sake of the ability to discriminate irrelevant paragraphs, irrelevant \textit{negative hop nodes} are added to $\&G$ in advance. As mentioned in \S~\ref{neg1}, the output of $[CLS]$, $\*T_0$, is in charge of generating \emph{negative threshold}. Therefore, $P_{ans}^{start}$ for each negative hop node is the one-hot vector where $\*{gt}_{ans}^{start}[0] = 1$.\label{neg} 

Cross entropy loss is used to train the span extraction task in System 1. The losses for the end position and for the next-hop spans are defined in the same way as follows. 
\begin{equation}
    \&L_{ans}^{start} = -\sum\limits_i \*{gt}_{ans}^{start}[i] \cdot \log P_{ans}^{start}[i]
\end{equation}
\subsubsection{Task \#2: Answer Node Prediction}
To command the reasoning ability, our model must learn to identify the correct answer node from a cognitive graph. For each question in the training set, we construct a training sample for this task. Each training sample is a composition of the \textit{gold-only graph}, which is the union of all correct reasoning paths, and negative nodes.  
Negative nodes include negative hop nodes used in Task \#1 and two negative answer nodes. A negative answer node is constructed from a span extracted at random from a randomly chosen hop node. 

For special question, we first compute the \textit{final answer probabilities} for each node by performing \textit{softmax} on the outputs of $\&F$. Loss $\&L$ is defined as cross entropy between the probabilities and one-hot vector of answer node $ans$. 
\begin{equation}
    \&L = -\log \Big(\textit{softmax}\big(\&F(\*X)\big)[ans]\Big) 
\end{equation}
Alternative and general questions are optimized by binary cross entropy in similar ways. 
The losses of this task not only are back-propagated to optimize predictors and System 2, but also fine-tune System 1 through semantic vectors $sem[x,Q,clues]$. 

\section{Experiment}\label{sec:exp}
\subsection{Dataset}\label{sec:dataset}
We use the full-wiki setting of HotpotQA to conduct our experiments. 
112,779 questions are collected by crowdsourcing based on the first paragraphs in Wikipedia documents, 84\% of which require multi-hop reasoning. The data are split into a training set (90,564 questions), a development set (7,405 questions) and 
a test set (7,405 questions).
All questions in development and test sets are \textit{hard multi-hop} cases.

In the training set, for each question, an answer and paragraphs of 2 \textit{gold} (useful) entities are provided, with multiple \textit{supporting facts}, sentences containing key information for reasoning, marked out. There are also 8 unhelpful \textit{negative paragraphs} for training. During evaluation, only questions are offered and meanwhile supporting facts are required besides the answer.  

To construct cognitive graphs for training, edges in gold-only cognitive graphs are inferred from supporting facts by fuzzy matching based on Levenshtein distance~\cite{navarro2001guided}.\label{fuzzy_match}
For each supporting fact in $para[x]$, if any gold entity or the answer, denoted as $y$, is fuzzy matched with a span in the supporting fact, edge $(x,y)$ is added.

\subsection{Experimental Details}
We use pre-trained BERT-base model released by \cite{devlin2018bert} in System 1. The hidden size $H$ is 768, unchanged in node vectors of GNN and predictors. All the activation functions in our model are \emph{gelu}~\cite{hendrycks2016bridging}. We train models on Task \#1 for 1 epoch and then on Task \#1 and \#2 jointly for 1 epoch. Hyperparameters in training are as follows: 
\begin{center}
\begin{small}
\setlength{\tabcolsep}{3pt}
\begin{tabular}{ccccc}
\hline
    Model&Task&batch size&learning rate&weight decay\\
    \hline
    BERT & \#1,\#2 & 10 & $10^{-4},4\times 10^{-5}$ & 0.01\\
    GNN & \#2 & graph & $10^{-4}$ & 0 \\
\hline
\end{tabular}
\end{small}
\end{center}

BERT and GNN are optimized by two different Adam optimizers, where $\beta_1 = 0.9, \beta_2 = 0.999$. The predictors share the same optimizer as GNN. The learning rate for parameters in BERT warmup over the first 10\% steps, and then linearly decays to zero.

To select out supporting facts, we just regard the sentences in the $clues$ of any node in graph as supporting facts. 
In the initialization of $\&G$, these 1-hop spans exist in the question and can also be detected by fuzzy matching with supporting facts in training set. The extracted 1-hop entities by our framework can improve the retrieval phase of other models (See \S~\ref{sec:baseline}), which motivated us to separate out the extraction of 1-hop entities to another BERT-base model for the purpose of reuse in implementation.
\subsection{Baselines}\label{sec:baseline}
\begin{table*}[]
\small{
    \centering
    \begin{tabular}{|c|c|cccc|cccc|cccc|c|}
    \hline
        \multirow{2}*{}&\multirow{2}*{Model} & \multicolumn{4}{|c|}{Ans} & \multicolumn{4}{|c|}{Sup} &         \multicolumn{4}{|c|}{Joint} \\
        \cline{3-14}
        ~ & ~& EM & $F_1$ & Prec & Recall & EM & $F_1$ & Prec & Recall & EM & $F_1$ & Prec & Recall\\
        \hline
        \multirow{5}*{Dev}&~\citet{yang2018hotpotqa} & 23.9 & 32.9 & 34.9 & 33.9 & 5.1 & 40.9 & 47.2 & 40.8 & 2.5 & 17.2 & 20.4 & 17.8\\
        ~&~\citet{yang2018hotpotqa}-IR&24.6 & 34.0 & 35.7 & 34.8 & 10.9 & 49.3 & 52.5 & 52.1 & 5.2 & 21.1 & 22.7 & 23.2\\
        ~&BERT& 22.7 & 31.6 & 33.4 & 31.9 & 6.5 & 42.4 & 54.6 & 38.7 & 3.1 & 17.8 & 24.3 & 16.2\\
        \cline{2-14}
        ~&\name-sys1&33.6 & 45.0 & 47.6 & 45.4 & \textbf{23.7} & 58.3 & \textbf{67.3} & 56.2 & 12.3 & 32.5 & 39.0 & 31.8\\
        ~&\name-onlyR&34.6 & 46.2 & 48.8 & 46.7 & 14.7 & 48.2 & 56.4 & 47.7 & 8.3 & 29.9 & 36.2 & 30.1\\
        ~&\name-onlyQ&30.7 & 40.4 & 42.9 & 40.7 & 23.4 & 49.9 & 56.5 & 48.5 & \textbf{12.4} & 30.1 & 35.2 & 29.9\\
        \cline{2-14}
        ~&\name&\textbf{37.6} & \textbf{49.4} & \textbf{52.2} & \textbf{49.9} & 23.1 & \textbf{58.5} & 64.3 & \textbf{59.7} & 12.2 & \textbf{35.3} & \textbf{40.3} & \textbf{36.5}\\
        
        \hline
        \multirow{4}*{Test}&~\citet{yang2018hotpotqa} & 24.0 & 32.9 & -  & - & 3.86 & 37.7 &-&-& 1.9 & 16.2&-&-\\
        ~&QFE & 28.7 & 38.1 & -  & - & 14.2 & 44.4 &-&-& 8.7 & 23.1&-&-\\
        ~&DecompRC & 30.0 & 40.7 & -  & - & N/A & N/A & - & - & N/A & N/A &-&-\\
        ~&MultiQA & 30.7 & 40.2 & - & - & N/A & N/A & - & - & N/A & N/A &-&-\\
        ~&GRN & 27.3 & 36.5 & - & - & 12.2 & 48.8 & - & - & 7.4 & 23.6 &-&-\\
        ~&\name& \textbf{37.1} & \textbf{48.9} & - &- & \textbf{22.8}&\textbf{57.7} & - & - & \textbf{12.4}& \textbf{34.9}& - & -\\
    \hline
    \end{tabular}
    }
    \addtolength{\belowcaptionskip}{-12pt}
    \caption{Results on HotpotQA (fullwiki setting). The test set is not public. The maintainer of HotpotQA only offers EM and $F_1$ for every submission. N/A means the model cannot find supporting facts.}
    \label{tab:result}
\end{table*}
The first category is previous work or competitor:
\begin{compactitem}
\setlength{\itemsep}{-12pt}
\item\textbf{\citet{yang2018hotpotqa}} The strong baseline model proposed in the original HotpotQA paper~\cite{yang2018hotpotqa}. It follows the retrieval-extraction framework of DrQA~\shortcite{chen2017reading} and subsumes the advanced techniques in QA, such as self-attention, character-level model, bi-attention.\\ 
\item\textbf{GRN}, \textbf{QFE}, \textbf{DecompRC}, \textbf{MultiQA}  The other models on the leaderboard.\footnote{All these models are unpublished before this paper.}\\ 
\item\textbf{BERT} State-of-art model on single-hop QA. BERT in original paper requires single-paragraph input and pre-trained BERT can barely handle paragraphs of at most 512 tokens, much fewer than the average length of concatenated paragraphs. We add relevant sentences from predecessor nodes in the cognitive graph to every paragraphs and report the answer span with maximum start probability in all paragraphs.\\
\item\textbf{\citet{yang2018hotpotqa}-IR\ } \citet{yang2018hotpotqa} with \textbf{I}mproved \textbf{R}etrieval. \citet{yang2018hotpotqa} uses traditional \textit{inverted index filtering} strategy to retrieve relevant paragraphs. The effectiveness might be challenged due to its failures to find out entities mentioned in question sometimes. The main reason is that word-level matching in retrieval usually neglect language models, which indicates importance and POS of words. We improve the retrieval by adding 1-hop entities spotted in the question by our model, increasing the coverage of supporting facts from 56\% to 72\%. 
\end{compactitem}
Another category is for ablation study:
\begin{compactitem}
\setlength{\itemsep}{-12pt}
\item\textbf{\name-onlyR} model initializes $\&G$ with the same entities retrieved in~\citet{yang2018hotpotqa} as 1-hop entities, mainly for fair comparison.\\ \item\textbf{\name-onlyQ} initializes $\&G$ only with 1-hop entities extracted from question, free of retrieved paragraphs. Complete \name~implementation uses both.\\
\item\textbf{\name-sys1} only retains \sysone~and lacks cascading reasoning in \systwo.
\end{compactitem}

\subsection{Results}
Following ~\citet{yang2018hotpotqa}, the evaluation of answer and supporting facts consists of two metrics: Exact Match (EM) and $F_1$ score. Joint EM is 1 only if answer string and supporting facts are both strictly correct. Joint precision and recall are the products of those of Ans and Sup, and then joint $F_1$ is calculated. All results of these metrics are averaged over the test set.\footnote{Thus it is possible that overall $F_1$ is lower than both precision and recall.} Experimental results show superiority of our method in multiple aspects:

\vpara{Overall Performance}Our \name~outperforms all baselines on all metrics by a significant margin (See Table~\ref{tab:result}). The leap of performance mainly results from the superiority of the \name~framework over traditional retrieval-extraction methods. 
Since paragraphs that are multi-hop away may share few common words literally or even little semantic relation with the question, retrieval-extraction framework fails to find the paragraphs that become related only after the reasoning clues connected to them are found.
Our framework, however, gradually discovers relevant entities following clues.


\vpara{Logical Rigor} QA systems are often criticized to answer questions with shallow pattern matching, not based on reasoning. To evaluate logical rigor of QA, we use $\frac{Joint EM}{Ans EM}$, the proportion of ``joint correct answers'' in correct answers. The joint correct answers are those deduced from all necessary and correct supporting facts. Thus, this proportion stands for logical rigor of reasoning. The proportion of our method is up to $33.4\%$, far outnumbering 7.9\% of ~\citet{yang2018hotpotqa} and $30.3\%$ of QFE.


\begin{figure}[hbt]
    \centering
    \includegraphics[width=\linewidth]{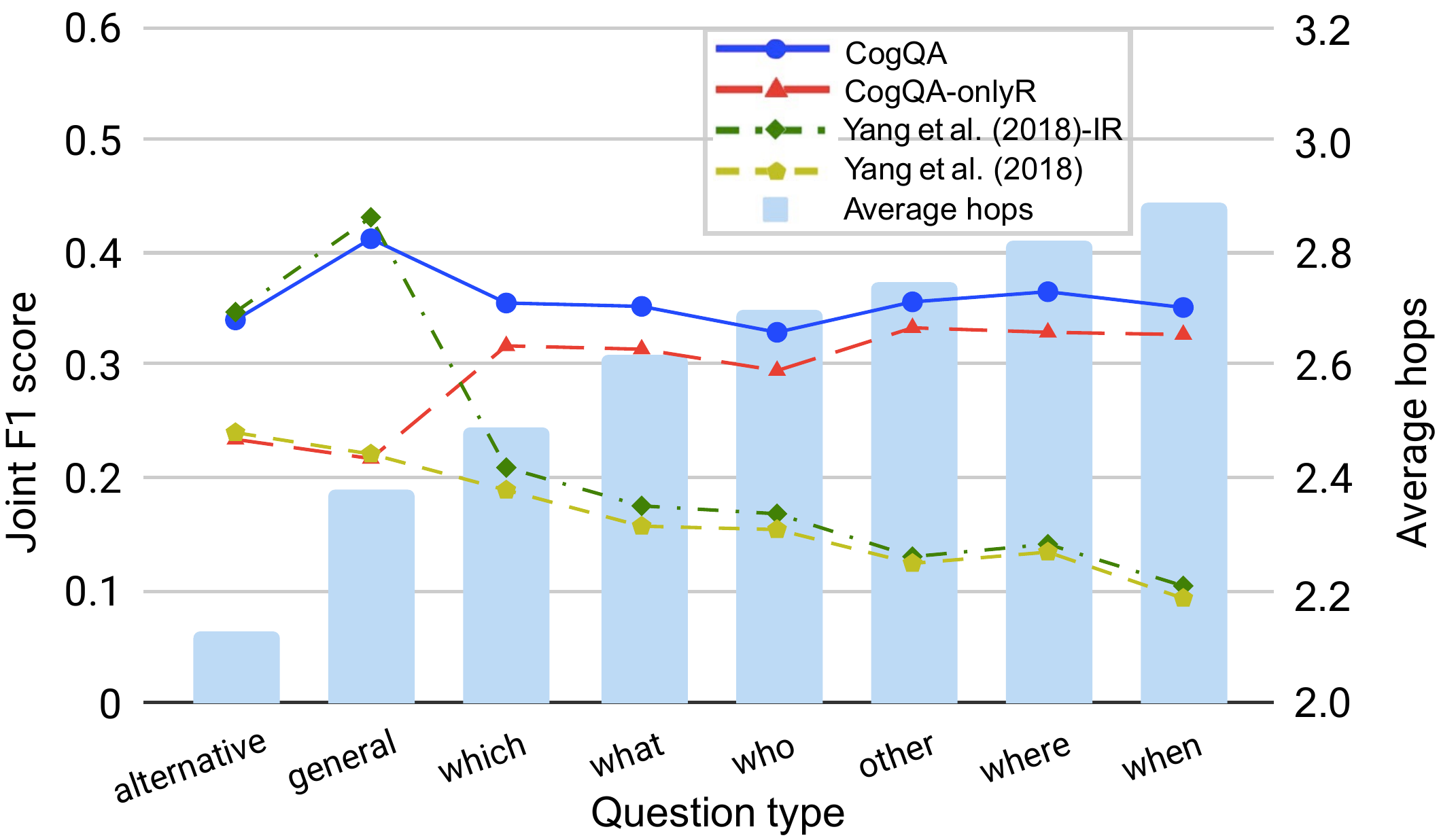}
    \addtolength{\belowcaptionskip}{-13pt}
    \addtolength{\abovecaptionskip}{-16pt}
    \caption{Model performance on 8 types of questions with different hops. 
    }
    \label{fig:question_types}
\end{figure}
\begin{figure*}[h]
    \centering
    \includegraphics[width=\textwidth]{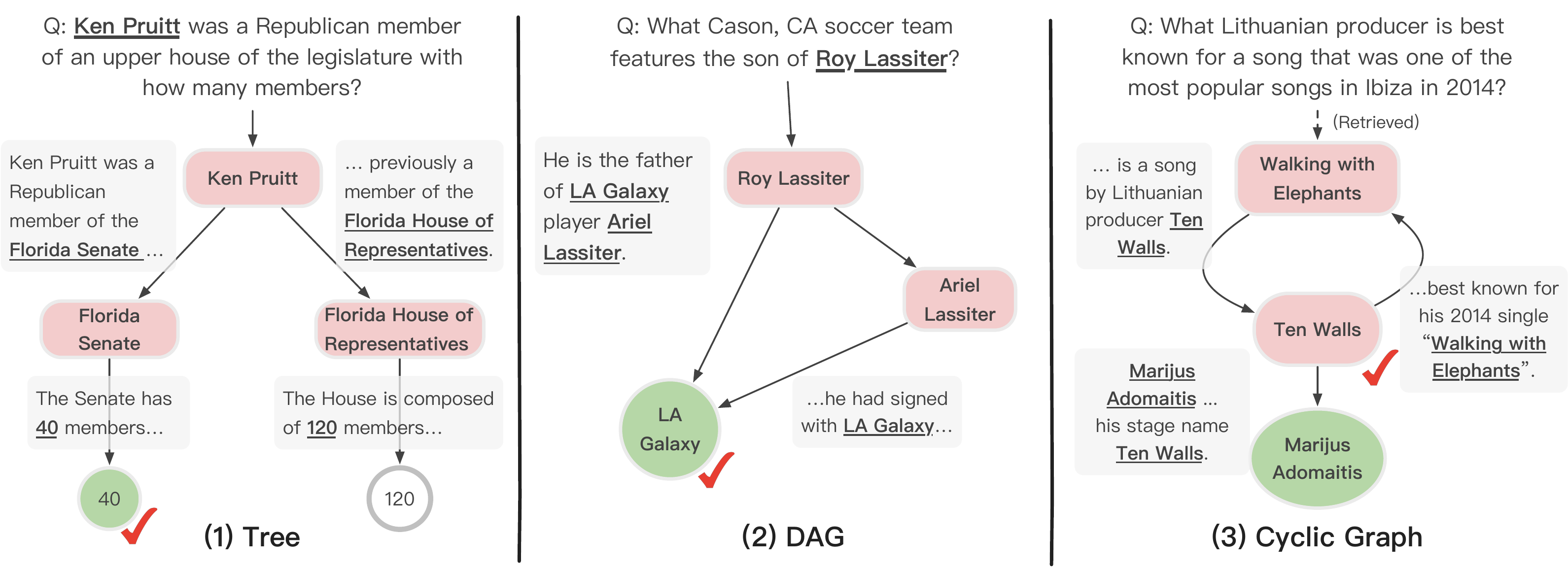}
    \addtolength{\belowcaptionskip}{-13pt}
    \addtolength{\abovecaptionskip}{-16pt}
    \caption{Case Study. Different forms of cognitive graphs in our results, i.e., Tree, Directed Acyclic Graph (DAG), Cyclic Graph. Circles are candidate answer nodes while rounded rectangles are hop nodes. Green circles are the final answers given by \name~and check marks represent the annotated ground truth.
    }
    \label{fig:case}
\end{figure*}
\vpara{Multi-hop Reasoning} Figure~\ref{fig:question_types} illustrates joint $F_1$ scores and average hops of 8 types of questions, including general, alternative and special questions with different interrogative word. 
As the hop number increases, the performance of ~\citet{yang2018hotpotqa} and ~\citet{yang2018hotpotqa}-IR drops dramatically, while our approach is surprisingly robust.
However, there is no improvement in alternative and general questions, because the evidence for judgment cannot be inferred from supporting facts, leading to lack of supervision. Further human labeling is needed to answer these questions.  

\vpara{Ablation Studies}
To study the impacts of initial entities in cognitive graphs, \name-onlyR begins with the same initial paragraphs as \cite{yang2018hotpotqa}. We find that \name-onlyR still performs significantly better. The performance decreases slightly compared to \name, indicating that the contribution mainly comes from the framework. 

To compare against the retrieval-extraction framework, \name-onlyQ is designed that it only starts with the entities that appear in the question. 
Free of elaborate retrieval methods, this setting can be regarded as a natural thinking pattern of human being, in which only explicit and reliable relations are needed in reasoning.  
\name-onlyQ still outperforms all the baselines, which may reveal the superiority of \name~framework over the retrieval-extraction framework.


BERT is not the key factor of improvement, although plays a necessary role. 
Vanilla BERT
performs similar or even slightly poorer to ~\cite{yang2018hotpotqa} in this multi-hop QA task, possibly because of the pertinently designed architectures in ~\citet{yang2018hotpotqa} to better leverage supervision of supporting facts.

To investigate the impacts of the absence of System 2, 
we design a System 1 only approach, \name-sys1, which inherits the iterative framework
but outputs answer spans with maximum predicted probability.
On \textit{Ans} metrics, the improvement over the best competitor decreases about 50\%, highlighting the reasoning capacity of GNN on cognitive graphs.

\vpara{Case Study}We show how the cognitive graph clearly explains complex reasoning processes in our experiments in Figure~\ref{fig:case}. The cognitive graph highlights the heart of the question in case (1) -- i.e., to choose between the number of members in two houses. \name~makes the right choice based on semantic similarity between ``Senate'' and ``upper house''. Case (2) illustrates that the robustness of the answer can be boosted by exploring parallel reasoning paths. Case (3) is a \emph{semantic retrieval} question without any entity mentioned, which is intractable for \name-onlyQ or even human. Once combined with information retrieval, our model finally gets the answer ``Marijus Adomaitis'' while the annotated ground truth is ``Ten Walls''. However, when backtracking the reasoning process in cognitive graph, we find that the model has already reached ``Ten Walls'' and answers with his real name, which is acceptable and even more accurate. Such explainable advantages are not enjoyed by black-box models.

\section{Related work}\label{sec:related}

\vpara{Machine Reading Comprehension} 
The research focus of machine reading comprehension (MRC) has been gradually transferred from cloze-style tasks~\cite{hermann2015teaching, hill2015goldilocks} to more complex QA tasks~\cite{rajpurkar2016squad} recent years. Compared to the traditional computational linguistic pipeline~\cite{hermann2015teaching}, neural network models, for example BiDAF~\cite{seo2016bidirectional} and R-net~\cite{wang2017gated}, exhibit outstanding capacity for answer extraction in text. Pre-trained on large corpra, recent BERT-based models nearly settle down the single paragraph MRC-QA problem with performances beyond human-level, driving researchers to pay more attention to multi-hop reasoning.

\vpara{Multi-Hop QA}
Pioneering datasets of multi-hop QA are either based on limited knowledge base schemas~\cite{talmor2018web}, or under multiple choices setting~\cite{welbl2018wikihop}. The noise in these datasets also restricted the development of multi-hop QA until high-quality HotpotQA~\cite{yang2018hotpotqa} is released recently. The idea of ``multi-step reasoning'' also breeds \textit{multi-turn} methods in single paragraph QA~\cite{kumar2016ask,seo2016query,shen2017reasonet}, assuming that models can capture information at deeper level implicitly by reading the text again.

\vpara{Open-Domain QA}
Open-Domain QA (QA at scale) refers to the setting where the search space of the supporting evidence is extremely large.
Approaches to get paragraph-level answers has been thoroughly investigated by the information retrieval community, which can be dated back to the 1990s~\citep{Belkin1993InteractionWT,voorhees1999trec,moldovan2000structure}. Recently, DrQA~\citep{chen2017reading} leverages a neural model to extract the accurate answer from retrieved paragraphs, usually called retrieval-extraction framework, greatly advancing this time-honored research topic again. Improvements are made to enhance retrieval by heuristic sampling~\cite{clark2017simple} or reinforcement learning~\cite{hu2017reinforced,wang2018r}, while for complex reasoning, necessary revisits to the framework are neglected.

\section{Discussion and Conclusion}\label{sec:conclusion}
We present a new framework \name~to tackle multi-hop machine reading problem at scale. The reasoning process is organized as cognitive graph, reaching unprecedented entity-level explainability. Our implementation based on BERT and GNN obtains state-of-art results on HotpotQA dataset, which shows the efficacy of our framework. 

Multiple future research directions may be envisioned. Benefiting from the explicit structure in the cognitive graph, \systwo~in \name~has potential to leverage neural logic techniques to improve reliability. Moreover, we expect that prospective architectures combining attention and recurrent mechanisms will largely improve the capacity of \sysone~by optimizing the interaction between systems. Finally, we believe that our framework can generalize to other cognitive tasks, such as conversational AI and sequential recommendation. 

\section*{Acknowledgements}
The work is supported by 
Development Program of China (2016QY01W0200),
NSFC for Distinguished Young Scholar (61825602),
NSFC (61836013), and a research fund supported by Alibaba.
The authors would like to thank Junyang Lin, Zhilin Yang and Fei Sun for their insightful feedback, and responsible reviewers of ACL 2019 for their valuable suggestions.
\bibliographystyle{acl_natbib}
\bibliography{references}
\end{document}